\documentclass[conference]{IEEEtran}
\IEEEoverridecommandlockouts

\usepackage{amsmath,amssymb,amsfonts}
\usepackage{algorithm, algorithmic}
\usepackage{textcomp}
\usepackage{mathtools}
\usepackage{xcolor}
\usepackage{microtype}
\usepackage{graphicx}
\usepackage[noadjust]{cite}
\usepackage{booktabs}
\usepackage{hyperref}

\def\BibTeX{{\rm B\kern-.05em{\sc i\kern-.025em b}\kern-.08em
    T\kern-.1667em\lower.7ex\hbox{E}\kern-.125emX}}

\makeatletter
\newcommand{\linebreakand}{%
\end{@IEEEauthorhalign}
\hfill\mbox{}\par
\mbox{}\hfill\begin{@IEEEauthorhalign}
}
\makeatother

\begin{document}

\title{LNPT: Label-free Network Pruning and Training\\}

\author{
	\IEEEauthorblockN{1\textsuperscript{st} Jinying Xiao}
	\IEEEauthorblockA{\textit{School of Computer and Communication Engineering} \\
		\textit{Changsha University of Science and Technology}\\
		Changsha, China \\
		xiaojinying1014@163.com}
	\\
		\IEEEauthorblockN{3\textsuperscript{rd} Zhe Tang}
	\IEEEauthorblockA{\textit{School of Computer and Communication Engineering} \\
		\textit{Changsha University of Science and Technology}\\
		Changsha, China \\
		tangzhe77777@163.com}
	
	\and
	\IEEEauthorblockN{2\textsuperscript{nd} Ping Li}
	\IEEEauthorblockA{\textit{School of Computer and Communication Engineering} \\
		\textit{Changsha University of Science and Technology}\\
		Changsha, China \\
		lping9188@163.com}
	\\
	\IEEEauthorblockN{4\textsuperscript{th} Jie Nie}
	\IEEEauthorblockA{\textit{School of Computer and Communication Engineering} \\
		\textit{Changsha University of Science and Technology}\\
		Changsha, China \\
		csustniejie@163.com}
}
\maketitle

\begin{abstract}
Pruning before training enables the deployment of neural networks on smart devices. By retaining weights conducive to generalization, pruned networks can be accommodated on resource-constrained smart devices. It is commonly held that the distance on weight norms between the initialized and the fully-trained networks correlates with generalization performance. However, as we have uncovered, inconsistency between this metric and generalization during training processes, which poses an obstacle to determine the pruned structures on smart devices in advance. In this paper, we introduce the concept of the learning gap, emphasizing its accurate correlation with generalization. Experiments show that the learning gap, in the form of feature maps from the penultimate layer of networks, aligns with variations of generalization performance. We propose a novel learning framework, LNPT, which enables mature networks on the cloud to provide online guidance for network pruning and learning on smart devices with unlabeled data. Our results demonstrate the superiority of this approach over supervised training.
\end{abstract}


\section{INTRODUCTION}
\noindent Recently, a large number of deep learning algorithms have been employed in smart devices. However, these methods are constrained by limited computational resources, resulting in challenges in deploying neural networks. Considering privacy and cost constraints, labels for training in specific scenarios are unknown. Our objective is to achieve adaptive deployment of neural networks on smart devices using unlabeled data. This issue stands as a focal point in both academia and industry \cite{ref3}\cite{ref4}\cite{ref5}.

Pruning enables neural networks to run on resource-constrained smart devices. Pre-pruning is usually the first choice for model installation before deployment of smart devices. According to the Lottery Ticket Hypothesis (LTH) \cite{ref9}, winning lottery ticket subnetworks exist within randomly initialized networks. LTH aims to preserve weights that are conducive to generalization as much as possible, making the measurement of generalization one of our focal points. Recent research\cite{ref14} has summarized metrics for the generalization performance of pruned networks, but this method primarily focuses on networks with different levels of sparsity. Therefore, explicit measurement of the network's generalization performance during the training process remains an area for further investigation.

On the other hand, a significant portion of network compression methods \cite{ref7}\cite{ref9}\cite{ref10}\cite{ref11}\cite{ref55} typically require labeled data for supervision during pruning or fine-tuning. These approachs not only compromise the privacy of the original data but also incur prohibitively high costs in collecting large-scale label-free data for many critical tasks, such as autonomous driving and healthcare \cite{ref41}. Therefore, the focus on label-free compression methods has been growing, significantly driving the development of various learning tasks, including objective detection and anti-face spoofing.

\begin{figure}[tb]
	\centering
	\includegraphics[width=0.4\textwidth, keepaspectratio]{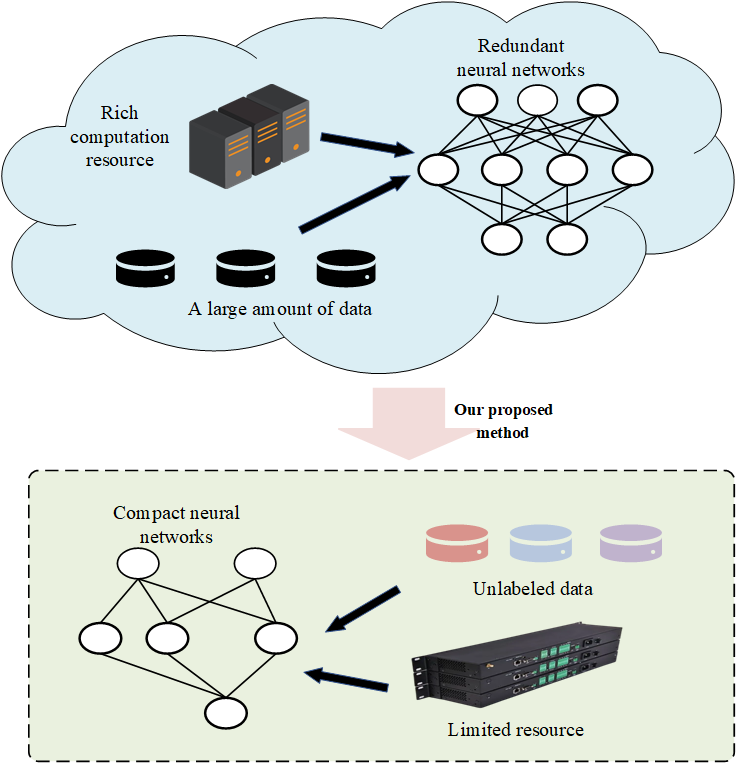}
	\caption{Our proposed method transforms cloud-scale large neural networks into lightweight neural networks for smart devices using unlabeled data.}
	\label{fig1.1}
\end{figure}

\begin{figure}[tb]
	\centering
	\includegraphics[width=0.45\textwidth, keepaspectratio]{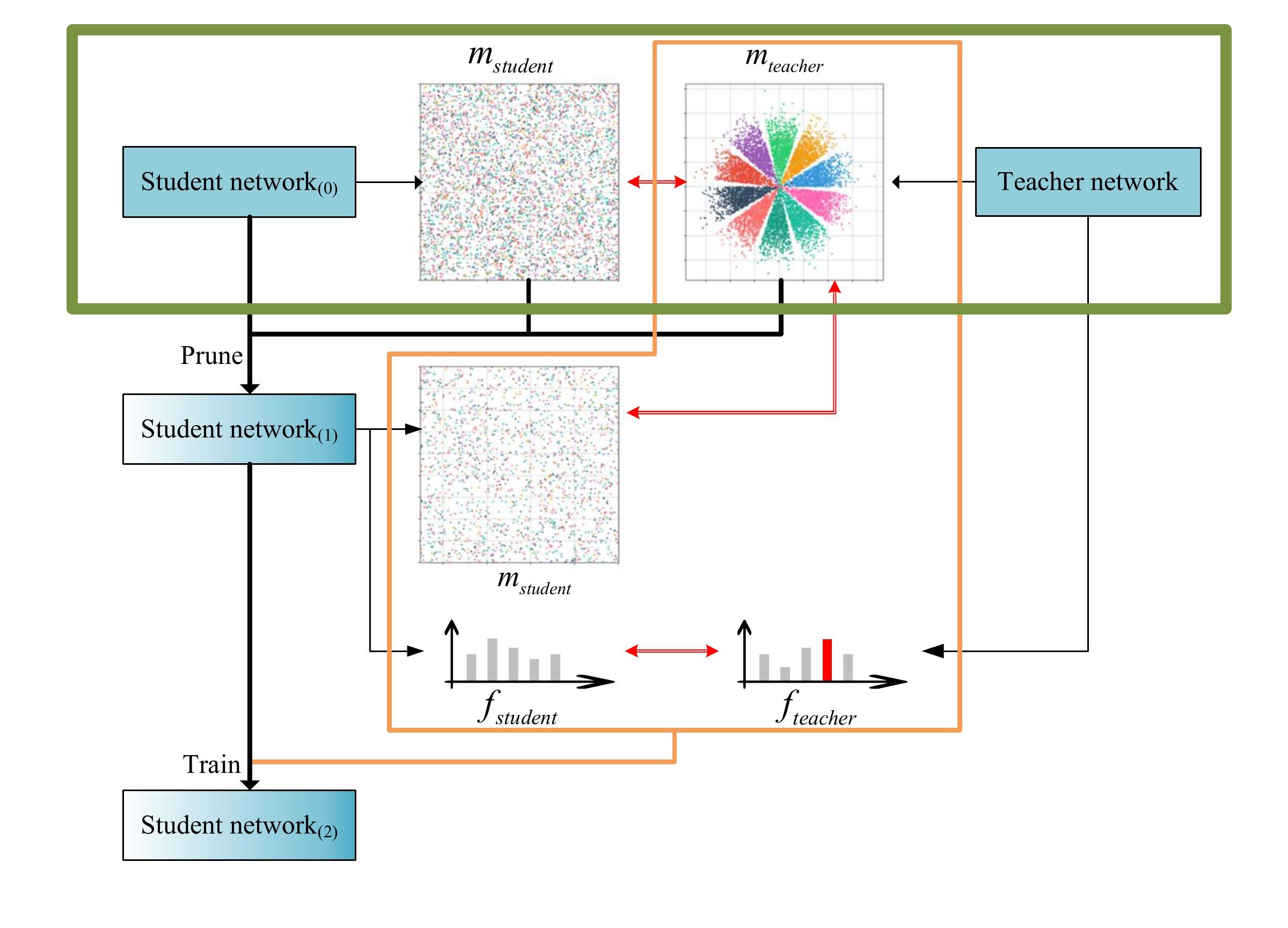}
	\caption{This figure provides an overview of the method described in the paper. In the figure, $\boldsymbol{m}$ represents the feature map in the penultimate layer, $f$ represents the network's output, and the subscripts "teacher" and "student" indicate whether the variable belongs to the teacher or the student network. 
		Within the "Student network" box, the subscripts (0), (1), and (2) indicate different states: (0) represents the initialized student, (1) represents the pruned student, and (2) represents the student after training. In the green box, it represents the student network undergoing pruning based on the guidance from the teacher's feature map. In the orange box, it shows the construction of a loss function based on the differences between feature maps and outputs to train the pruned student.}
	\label{fig1}
\end{figure}

To address the aforementioned issues, a characterization of generalization performance is proposed by considering the gap between the feature maps in the penultimate layer of the network (hereinafter referred to as the feature map). Based on this feature map gap, we introduce a novel pruning-train framework, LNPT, consisting of two steps: 1) pruning the student, i.e., the target network. 2) training the student using the teacher's outputs and feature maps. As shown in Fig.\ref{fig1.1}, our approach allows for guidance of a small student network by a cloud-based teacher network, and training on smart devices does not require labels. It is worth noting that our approach does not require labels compared to other pruning methods. The detailed steps of our method are illustrated in Fig.\ref{fig1}.

Our main contributions are summarized as follows:
\begin{itemize}
	\item We observe that the changes in the distance of weight norms and generation are inconsistent during training with regard to various pruning rates. It seems that the distance is to escape constraints from the expected training trend, a phenomenon we called weight escape. As weight escape make it unreasonable to train models after pruning, we introduce the concept of ’learning gap’ to describe the correlation between network weight norms and generalization. This not only characterizes the process of enhancing generalization but also serves as a prerequisite for pre-pruning.
	\item We argue that the reason for weight escape phenomenon is due to intense variability of loss gradient. Due to the unique properties of feature maps and our discoveries during this work, a synaptic is constructed based on feature maps to serve as the pruning criterion. Compared to the typical fitting loss construction methods, this criterion is less influenced by the individual features of the data and exhibits smoother variations. Experimental validation confirms that this criterion adheres closely to the nearly consistent generalization.
	\item We introduce LNPT, a method capable of efficient pruning and training of students without labeled data. The superiority of this method is demonstrated on CIFAR10/100 and Tiny-Imagenet datasets, with results that even surpass labeled methods.
\end{itemize}

\section{RELATED WORK}

\subsection{Pruning Based on Feature Maps}

\noindent The earliest pruning algorithms considered removing the least impactful weights based on pre-trained networks \cite{ref6}. Later, methods based on magnitude pruning followed by iterative fine-tuning and retraining achieved good performance. Various methods have been defined to assess the importance of weights (filters) for the removal of redundant parameters. These methods include evaluations like Fast Student \cite{ref27}, weight sensitivity to the loss function \cite{ref10}, and gradient flow sensitivity \cite{ref11}.

In feature map-based pruning methods, some approaches proposed using both class information from network outputs and intermediate feature information to score filters \cite{ref25}\cite{ref26}. These methods assessed the contribution of filters based on the rank information of feature maps. While these criteria based on different attributes have achieved success in network compression, they may suffer performance degradation under certain conditions. This is because a single criterion is often limited in considering all factors such as feature size or feature dimension simultaneously.
\subsection{Label-free distillation}
\noindent Since the original paper by Hinton \cite{ref8}, several subsequent works \cite{ref16}\cite{ref17} have developed advanced distillation techniques aimed at strengthening the consistency between teachers and students.

In particular, for label-free distillation, a significant body of knowledge distillation (KD) research has focused on leveraging logit information. \cite{ref18} introduced an additional small-scale auxiliary network to narrow the gap between the teacher and student. \cite{ref19} decoupled the classical KD loss, making the distillation loss more effective and flexible. \cite{ref31} transformed the teacher's logits into pseudo ground-truth labels for the student to learn from. However, introducing a teacher classifier can increase deployment costs and may contradict the principles of knowledge distillation. Moreover, the performance gap issue is far from being resolved.

To alleviate the performance gap between teachers and students in logit distillation, some work leverages feature information embedded in intermediate layers, aiming to compel students to mimic the teacher's feature representations. Feature distillation was initially proposed in FitNet \cite{ref23}, which employed "hints" to make the student's intermediate features mimic the corresponding parts of the teacher. Inspired by FitNet, various methods have been proposed to match these features. Specifically, attention maps of the teacher network \cite{ref20}, neural selectivity \cite{ref21}, and semantic information \cite{ref22} have been introduced to express knowledge. Feature-based distillation extracts richer information, providing greater flexibility for knowledge transfer. However, due to differences in feature sizes between teacher and student networks, feature-based distillation requires additional layer transformations to align different sizes.
\section{METHOD}
\label{section method}

\subsection{Notations}
\noindent To facilitate problem description, we define the network as ${\boldsymbol{\theta}}$, where $\boldsymbol{\theta}\in\mathbb{R}^{P\times1}$. For the feature map outputs of the teacher network $\boldsymbol{\theta}_t$ and the student network $\boldsymbol{\theta}_s$, we define them as $map_t$ and $map_s$, which we will refer to as $\boldsymbol{m}_t$ and $\boldsymbol{m}_s$, where $\boldsymbol{m}_t,\boldsymbol{m}_s\in\mathbb{R}^{M\times1}$.

\subsection{Learning Gap}
\label{subsection lg}
\noindent The norm values of weights capture feature information from the data during the training process. \cite{ref14} indicates that for pruned networks, the distance $\boldsymbol{d}\in\mathbb{R}^{P\times1}$ between the initialized network and the fully-trained network represents the generalization performance to some extent. This approach enables the exploration of network generalization through structural insights. Additionally, recent research \cite{ref28}\cite{ref29} has discovered that learning distance is closely related to the generalization of deep learning.

\begin{figure}[tb]
	\centering
	\includegraphics[width=0.4\textwidth, keepaspectratio]{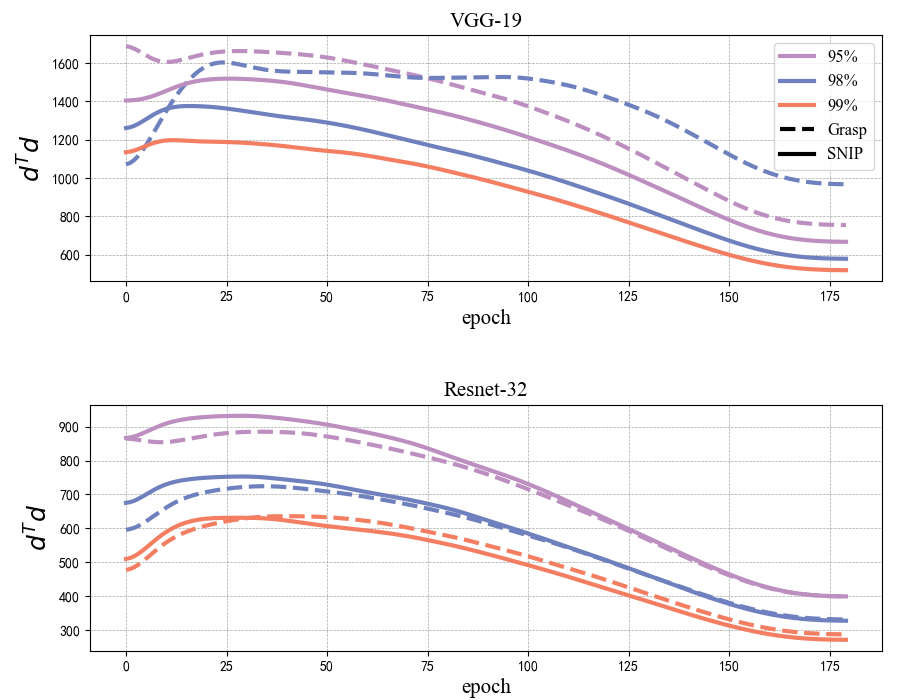}
	\caption{The figure illustrates the change in the teacher-student distance during the training phase for different algorithms and compression rates on the CIFAR-10 dataset. The percentages indicate the pruning rates, and the last batch within each epoch is sampled to measure the teacher-student distance $\boldsymbol{d}^T\boldsymbol{d}$. }
	\label{fig2}
\end{figure}

Fig.\ref{fig2} depicts the evolution of teacher-student distance during training. As observed in the figure, $\boldsymbol{d}^T\boldsymbol{d}$ does not exhibit the expected stable and consistent descent. Instead, it rises during the early stages of training, forming a convex trend. This phenomenon makes it challenging to establish a consistent correlation between teacher-student distance and generalization. We refer to this phenomenon as weight escape.

The emergence of the weight escape phenomenon has cast doubt on the scientific validity of pre-pruning as the provided pruning criteria before training is unable to predict subsequent changes. It is necessary to explore a more reasonable expression that more consistently correlates $\boldsymbol{d}$ with generalization. We propose a more expressive learning gap, denoted as:
\begin{equation}
	\label{eq1}
	\boldsymbol{\ell}=\boldsymbol{C}^t\boldsymbol{d}=\boldsymbol{C}^t\left(\boldsymbol{\theta}_t-\boldsymbol{\theta}_s\right)
\end{equation}

Where $\boldsymbol{C}^t$ represents the coefficient matrix at time t. $\boldsymbol{C}^t$ aligns the gradient fit with the variations in $\boldsymbol{\ell}$, enabling them to move in harmony. In the training process with data fitting as the objective, the change in $\boldsymbol{d}$ can be expressed as:
\begin{equation}
	\label{eq2}
	\Delta\boldsymbol{d}=\eta\nabla_{\boldsymbol{\theta}_s}Loss
\end{equation}
where $\eta$ represents the learning rate, Loss denotes the fitting loss, and $\nabla_{\boldsymbol{\theta}_s}$Loss represents the fitting gradient. Therefore, concerning the change in $\boldsymbol{\ell}$:
\begin{equation}
	\label{eq3}
	\Delta\boldsymbol{\ell}=\eta \boldsymbol{C}^t\nabla_{\boldsymbol{\theta}_s}Loss
\end{equation}

We aim for the transformation through $\boldsymbol{C}^t$ to make the variations in $\boldsymbol{C}^t\nabla_{\boldsymbol{\theta}_s}Loss$ stabilize during different training stages. Concerning the learning gap $\boldsymbol{\ell}$, if it steadily decreases as training iterations progress, the network will converge faster. This aligns with the observation in \cite{ref9} that "pruned networks learn faster and achieve higher test accuracy and better generalization than the original network." Therefore, through the coefficient matrix $\boldsymbol{C}^t$, we provide an equivalent expression for the relationship between $\boldsymbol{\ell}$ and network generalization, and the presence of $\boldsymbol{C}^t$ supports the feasibility of pre-pruning.

However, solving the coefficient matrix for large-scale networks is computationally infeasible. An alternative method that can closely reflect the change $\Delta\boldsymbol{\ell}$ in $\boldsymbol{\ell}$ during the training process needs to be proposed. It's worth noting that the fitting gradient is significantly affected by the individual features of batches, leading to increased network overfitting and significant fluctuations. Therefore, it often requires adding penalties to suppress the individual features of the fitting gradient \cite{ref43}, and regularization of the network is a primary means of improving generalization performance. This implies that constructing an expression for $\Delta\ell$ based on the fitting gradient is not reasonable.

\subsection{Pruning Criterion}
\noindent As we have addressed, it is not feasible to compute the learning gap straightforward. However, (\ref{eq1}) sheds light on requisite for the establishment of the pruning criteria before training. In other words, there exists a certain degree of similarity on intrinsic mechanism for changes between the expected criteria and weight norms. It is not reasonable to construct pruning criteria with data-fitting based gradient due to its strong dynamics during training.

In neural networks, feature maps can capture a wealth of information \cite{ref15}. Unlike \cite{ref25}\cite{ref26}, the student network learns by matching feature maps, encouraging the network to learn shared semantics between feature mappings, which is more conducive to student network learning \cite{ref30}. Thus, an attempt is made to use feature map-based gradients to construct the pruning criterion.

For the same batch of data $\left(x^{\left(k\right)},y^{\left(k\right)}\right)$ from a dataset $\{\left(x^{\left(k\right)},y^{\left(k\right)}\right)\}_{k=1}^N$, the teacher's feature map $\boldsymbol{m}_t$ guides the student network, and the student's feature map $\boldsymbol{m}_s$ approaches the teacher's feature map $\boldsymbol{m}_t$:
\begin{equation}
	\label{eq4}
	\Delta \boldsymbol{m}=\boldsymbol{m}_t-\boldsymbol{m}_s
\end{equation}

Therefore, we define the loss as:
\begin{equation}
	\label{eq5}
	L_{\boldsymbol{m}}\left(\boldsymbol{\theta}_s\right)=\Delta \boldsymbol{m}^T\Delta \boldsymbol{m}
\end{equation}

For a specific weight $\boldsymbol{\theta}_s^{\left(i\right)}$, the change in feature map loss, denoted as $\Delta L_{\boldsymbol{m}}$, is defined with the following expression:
\begin{equation}
	\label{eq6}
	\begin{split}
		\Delta L_{\boldsymbol{m}}&=\lim_{\varepsilon\to0}\frac{L_{\boldsymbol{m}}\left(\boldsymbol{\theta}_s^{\left(i\right)}+\varepsilon\frac{\partial L_{\boldsymbol{m}}\left(\boldsymbol{\theta}_s\right)}{\partial\boldsymbol{\theta}_s^{\left(i\right)}}\right)-L_{\boldsymbol{m}}\left(\boldsymbol{\theta}_s\right)}{\varepsilon}\\
		&=\frac{\partial L_{\boldsymbol{m}}\left(\boldsymbol{\theta}_s\right)}{\partial\boldsymbol{\theta}_s^{\left(i\right)}}\frac{\partial L_{\boldsymbol{m}}\left(\boldsymbol{\theta}_s\right)}{\partial\boldsymbol{\theta}_s^{\left(i\right)}}
	\end{split}
\end{equation}

Then, we introduce the perturbation $\delta$ to describe the feature gradient flow during training and use the Taylor approximation to describe how the gradient flow changes when this weight $\boldsymbol{\theta}_s^{\left(i\right)}$ is removed:

\begin{equation}
	\label{eq7}
	\begin{split}
		\Delta L_{\boldsymbol{m}}\left(\boldsymbol{\theta}_s^{\left(i\right)}+\delta\right)-\Delta L_{\boldsymbol{m}}\left(\boldsymbol{\theta}_s^{\left(i\right)}\right)\\=2\delta\frac{\partial^2L_{\boldsymbol{m}}\left(\boldsymbol{\theta}_s\right)}{\partial\left(\boldsymbol{\theta}_s^{\left(i\right)}\right)^2}\frac{\partial L_{\boldsymbol{m}}\left(\boldsymbol{\theta}_s\right)}{\partial\boldsymbol{\theta}_s^{\left(i\right)}}
	\end{split}
\end{equation}

Then, based on the scoring form of synaptic\cite{ref42}, the scoring function is given by:
\begin{equation}
	\label{eq8}
	S\left(\boldsymbol{\theta}_s^{\left(i\right)}\right)=\left|\boldsymbol{\theta}_s^{\left(i\right)}\frac{\partial^2L_{\boldsymbol{m}}\left(\boldsymbol{\theta}_s\right)}{\partial\left(\boldsymbol{\theta}_s^{\left(i\right)}\right)^2}\frac{\partial L_{\boldsymbol{m}}\left(\boldsymbol{\theta}_s\right)}{\partial\boldsymbol{\theta}_s^{\left(i\right)}}\right|
\end{equation}

A pruning mask is obtained for a given pruning ratio $\mathcal{P}$ by calculating scores for each weight and removing the top $\mathcal{P}$ portion of weights (see Algorithm \ref{alorithm_LNPT}).

\subsection{Knowledge Distillation Based on the Feature Maps}
In a K-class classification problem, for a batch of data with N samples $\left(x^{\left(i\right)},y^{\left(i\right)}\right)$, neural networks are typically trained by minimizing the cross-entropy loss $\mathcal{H}_{cross}$ over the entire dataset. KD modifies the objective by adding another regularization term instead of optimizing $\mathcal{H}_{cross}$, as follows:
\begin{equation}
	\label{eq9}
	\begin{split}
		\mathcal{L}_{KD}=\mathcal{H}_{cross}\left(\sigma\left(f\left(\boldsymbol{\theta},x^{\left(i\right)}\right)\right),y^{\left(i\right)}\right)\\+\alpha\mathcal{H}_{cross}\left(\sigma\left(f\left(\boldsymbol{\theta},x^{\left(i\right)}\right)\right),\sigma\left(f_t\left(\boldsymbol{\theta}_t,x^{\left(i\right)}\right)/T\right)\right)
	\end{split}
\end{equation}

Where $f_t\left(\theta_t,x^{\left(i\right)}\right)$ is the output of the teacher network, $\sigma(\cdot)$ is the softmax normalization function, $\alpha$ controls the balance between the two terms, and the temperature T, introduced in \cite{ref8}, acts as a factor for smoothing the teacher and student outputs.

However, due to privacy concerns, the labels cannot be accessed. Therefore, the first term in (\ref{eq9}) cannot be constructed. In \cite{ref31}, the output of the teacher network $\sigma\left(f_t\left(\theta_t,x^{\left(i\right)}\right)\right)=\left\{f_1,f_2\ldots f_N\right\},f_n\in\mathbb{R}^K$ is converted into one-hot vectors $t^{\left(i\right)}=\left\{t_1,t_2\ldots t_N\right\},t_n\in\mathbb{R}^K$, which serve as pseudo ground-truth labels, i.e.:
\begin{equation}
	\mathcal{L}_{oh}=\mathcal{H}_{cross}\left(\sigma\left(f\left(\boldsymbol{\theta},x^{\left(i\right)}\right)\right),t^{\left(i\right)}\right)
\end{equation}
Where:$$
t_n^j= \left\{
\begin{array}{l}
	1,j=\mathop{\arg\max}\limits_{i}f_n^i \\
	0,j\ne\mathop{\arg\max}\limits_{i}f_n^i
\end{array}
\right.
$$

In addition to hard labels, the introduction of soft labels brings the probability similarities and differences that are essential for the network. \cite{ref12} indicates that KD provides a more accurate estimate of label noise, allowing KD to better approximate the true distribution and thereby enhance the student's performance.

The difference between the teacher and student feature maps is considered indicative of the student's generalization performance. Therefore, we define the loss as:
\begin{equation}
	\label{eq11}
	\mathcal{L}_{\boldsymbol{m}}=MSE\left(\boldsymbol{m}_t,\boldsymbol{m}_s/T\right)
\end{equation}

Here, $MSE(\cdot)$ represents mean squared error, and T is a hyperparameter. Since larger feature map norms result in outputs closer to one-hot vectors \cite{ref32}, A temperature coefficient T is introduced for feature maps with excessively large individual norms. This approach allows the student to learn both the true labels through $\mathcal{L}_{oh}$ and move closer to the teacher through smoothed feature maps. Therefore, we define the total loss as:
\begin{equation}
	\label{eq12}
	\mathcal{L}_{total}=\mathcal{L}_{oh}+\alpha\mathcal{L}_{\boldsymbol{m}}
\end{equation}

Where $\alpha$ is a hyperparameter that balances the two terms.

Algorithm \ref{alorithm_LNPT} summarizes the detailed steps of feature map distillation.

\subsection{Dynamism of Feature Maps}
Recent works \cite{ref33}\cite{ref34} have investigated the training dynamics of large deep neural networks under gradient descent. Let $\mathcal{X}$,$\mathcal{Y}$ represent the inputs and labels, respectively. Consider a neural network $f(\cdot;\boldsymbol{\theta})$ parameterized by weights $\boldsymbol{\theta}$ and let $f\left(\mathcal{X};\boldsymbol{\theta}\right)$ be the concatenated vector of predictions for all training inputs. We emphasize the time-dependence of $\boldsymbol{\theta}$, where the NTK at time t is defined as \cite{ref35}:
\begin{equation}
	\label{eq13}
	\mathbf{\Theta}_\mathbf{t}\left(\mathcal{X},\mathcal{X};\boldsymbol{\theta}_s^t\right)=\nabla_{\boldsymbol{\theta}_s^t}f\left(\mathcal{X};\boldsymbol{\theta}_s^t\right)^T\nabla_{\boldsymbol{\theta}_s^t}f\left(\mathcal{X};\boldsymbol{\theta}_s^t\right)
\end{equation}

Where $\boldsymbol{\theta}_s^t$ represents the weights of the student network at time t.

In our approach, the aim is for the student network to pay more attention to the dynamics of feature maps between the teacher and the student in order to achieve efficient pruning results. In the student network, the last layer is a fully connected operation that takes the feature map $\boldsymbol{m}_s$ as input and produces logits $f_s$ through a linear transformation $f_s=W\boldsymbol{m}_s$. We define our optimization target as $\Delta_{\boldsymbol{m}}\left({\mathcal{X},\boldsymbol{\theta}}_s\right)=\boldsymbol{m}_t-\boldsymbol{m}_s$. As a result, for the weights of the layers related to feature maps, their sensitivity to the weights at time t is expressed in terms of gradient norm, as shown in (\ref{eq4}). Therefore, it can be formulated as:
\begin{equation}
	\label{eq14}
	{s\left(\boldsymbol{\theta}_s^t\right)=W^+\mathbf{\Theta}\left(\mathcal{X},\mathcal{X};\boldsymbol{\theta}_s^t\right)\left(W^+\right)}^T
\end{equation}

Where$W^+$ is the Moore-Penrose pseudoinverse of $W$.

This indicates that the sensitivity of the weights to the target $\Delta_m\left({\mathcal{X},\theta}_s\right)$ leads to an increase in $\mathbf{\Theta}$ in the corresponding gradient direction. This suggests that the student can efficiently prune at the initial stage by capturing the dynamic sensitivity between the teacher and student, and \cite{ref35} has already demonstrated that this kernel can capture training dynamics throughout the entire training process.Therefore, we can conclude:
\begin{equation}
	s\left(\boldsymbol{\theta}_s^{t=0}\right)=s\left(\boldsymbol{\theta}_s^t\right)
\end{equation}

So, the optimization objective we propose can achieve efficient optimization throughout the training process.

\begin{algorithm}[tb]
	\caption{LNPT}
	\label{alorithm_LNPT}
	\begin{algorithmic}[1]
		\REQUIRE pruning ratio $\mathcal{P}$, pruning mask $\mathcal{M}$,training data $\mathcal{D}$, output of teacher network $f\left(\boldsymbol{\theta}_t\right)$ and feature map $\boldsymbol{m}_t$, output of student network $f\left(\boldsymbol{\theta}_s\right)$, feature map of student $\boldsymbol{m}_{s}$, number of main training steps $N$, learning rate $\alpha$
		
		\STATE $\mathcal{D}_b=\left\{\left(x_i,y_i\right)\right\}_{i=1}^b~\mathcal{D}$
		\STATE Compute the scores $S\left(\boldsymbol{\theta}_s\right)$(see (\ref{eq8})) 
		\STATE Compute $\mathcal{P}_{th}$ percentile of $S\left(\boldsymbol{\theta}_s\right)$ as $\tau$
		\STATE $\mathcal{M}=S\left(\boldsymbol{\theta}_s\right)<\tau$
		\STATE $\mathcal{L}_{total}=\mathcal{L}_{oh}+\alpha\mathcal{L}_{\boldsymbol{m}}$
		\vspace{0.1cm}
		\hrule
		\vspace{0.1cm}
		\FOR{$i$ from 1 to $N$}
		\STATE $\boldsymbol{\theta}_s^{i+1}=\boldsymbol{\theta}_s^i-\alpha\nabla_{\boldsymbol{\theta}_s}\mathcal{L}_{total}$
		\ENDFOR
	\end{algorithmic}
\end{algorithm}

\section{EXPERIMENT}
\noindent We conducted empirical evaluations of our LNPT method by comparing it with various state-of-the-art visual classification baselines on different architectures and datasets. In the context of image classification tasks, we explored several recent training-before-pruning schemes, including both unstructured pruning methods such as one-shot pruning (SNIP) \cite{ref10} and GraSP \cite{ref11}, as well as structured pruning methods like SSS \cite{ref39}, GAL \cite{ref40}, and HRank \cite{ref25}.In order to obtain more stable results, each experiment was conducted five times, and the results were averaged.
\begin{table}[tb]
	\centering
	\caption{Performance comparison of pruned VGG19 and ResNet32 on CIFAR10/100. The bolded number is the one with the highest accuracy among SNIP, GraSP, and LNPT. The \# indicates that the network failed to converge.}
	\label{table1}
	\resizebox{1.0\columnwidth}{!}{
		\begin{tabular}{l ccc ccc} 
			\cmidrule[\lightrulewidth](lr){1-7}
			\textbf{Dataset} & \multicolumn{3}{c}{CIFAR10} & \multicolumn{3}{c}{CIFAR100} \\ 
			\cmidrule(lr){1-7}
			\textbf{VGG19} & \multicolumn{3}{c}{Acc: 94.20\%} & \multicolumn{3}{c}{Acc: 74.16\%} \\ 
			\cmidrule(lr){1-7}
			Pruning ratio & 95\% & 98\% & 99\% & 95\% & 98\% & 99\% \\ 
			\cmidrule(lr){1-1} \cmidrule(lr){2-4} \cmidrule(lr){5-7}
			SNIP  &	
			\textbf{93.71} & 92.22 & \# &
			\textbf{71.95} & 56.23 & \# \\
			GraSP  &	
			93.07 & 92.26 & 91.57 &	
			71.62 & 70.15 & 66.72 \\
			SSC\cite{ref49} &
			93.45 & 91.1 & - &	
			70.71 & 68.95 & - \\
			SynFlow &
			93.37 & 92.24& 91.04 &	
			71.18 & 67.93 & 64.95 \\
			IP-SynFlow\cite{ref50} &
			92.46 & 92.11& - &	
			- & - & - \\
			LNPT &
			93.41 & \textbf{92.62} & \textbf{92.06} &	
			71.49 & \textbf{70.52} & \textbf{67.33} \\

	\end{tabular}}
	
	\resizebox{1.0\columnwidth}{!}{
		\begin{tabular}{l ccc ccc}
			\cmidrule[\lightrulewidth](lr){1-7}
			\textbf{ResNet32} & \multicolumn{3}{c}{Acc: 94.80\%} & \multicolumn{3}{c}{Acc: 74.83\%} \\ 
			\cmidrule(lr){1-7}
			Pruning ratio & 95\% & 98\% & 99\% & 95\% & 98\% & 99\% \\ 
			\cmidrule(lr){1-1} \cmidrule(lr){2-4} \cmidrule(lr){5-7}
			SNIP  &	
			91.51 & 88.36 & 85.12 & 65.03 & 52.19 & 36.04 \\
			GraSP  &	
			91.77 & 89.53 & 85.52 & 66.95 & 58.55 & 49.10 \\
			Deep-R\cite{ref51}  &	
			89.94 & 86.45 & - & 63.90 & 58.47 & - \\
			SSC\cite{ref49}  &	
			91.61 & 88.25 & - & 66.44 & 58.79 & - \\
			LNPT &
			\textbf{91.85} & \textbf{89.87} & \textbf{85.63} & 
			\textbf{67.14} & \textbf{59.15} & \textbf{51.44}\\
			\cmidrule[\lightrulewidth](lr){1-7}
	\end{tabular}}
\end{table}
\begin{table}[tb]
	\centering
	\caption{Performance comparison of pruned VGG19 and ResNet32 on Tiny-ImageNet.}
	\label{table2}
	\resizebox{1.0\columnwidth}{!}{
		\begin{tabular}{l ccc ccc}
			\cmidrule[\heavyrulewidth](lr){1-7}
			\textbf{Network} & \multicolumn{3}{c}{VGG19: 63.29\%} & \multicolumn{3}{c}{ResNet32: 63.86\%} \\ 
			\cmidrule(lr){1-1} \cmidrule(lr){2-4} \cmidrule(lr){5-7}
			Pruning ratio & 95\% & 98\% & 99\% & 95\% & 98\% & 99\% \\ 
			\cmidrule(lr){1-1} \cmidrule(lr){2-4} \cmidrule(lr){5-7}
			SNIP &	
			59.27 &	48.95 &	\# &	40.41 &	24.81 & 18.58 \\
			GraSP &	
			59.53 &	\textbf{56.54} &	52.36 & 48.45 &	37.25 &	26.17 \\
			LNPT &
			\textbf{59.84} & 55.91 & \textbf{53.21} &
			\textbf{48.64} & \textbf{37.96} & \textbf{27.35}  \\
			\midrule
	\end{tabular}}
\end{table}
\subsection{Unstructured Pruning}
\begin{figure}[tb]
	\centering
	\includegraphics[width=0.35\textwidth, keepaspectratio]{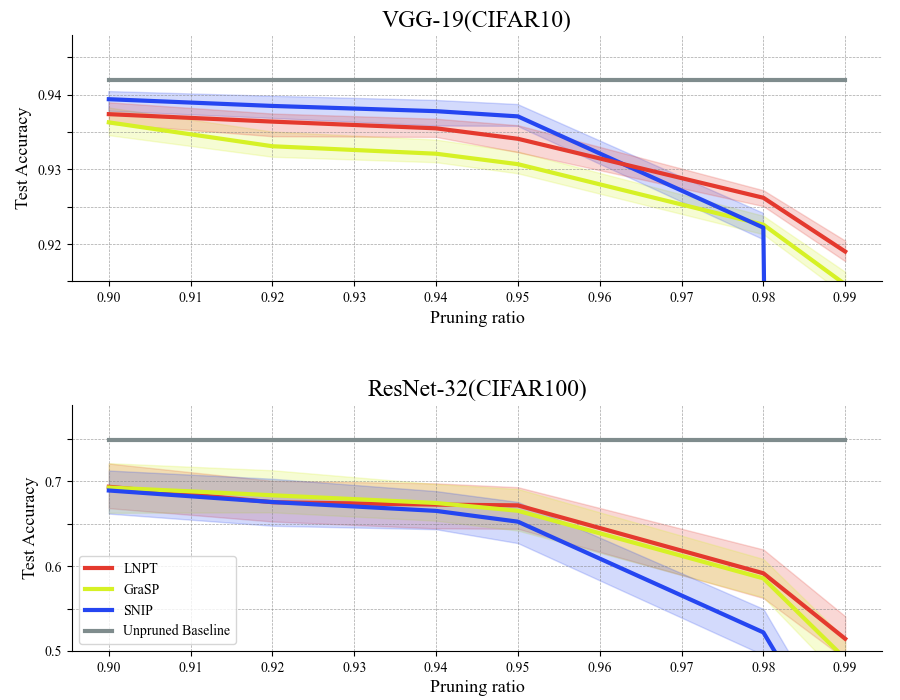}
	\caption{The accuracy of LNPT compared to other pre-pruning methods. The shaded area represents the range of deviations observed in repeated experiments.}
	\label{fig3}
\end{figure}

Two network architectures, VGG and ResNet, were used to assess LNPT's performance on image classification tasks across various pruning ratios. The entire network with L2 regularization constraint served as the baseline. LNPT single-shot pruning efficiency was evaluated on VGG19 and ResNet32 networks using datasets such as CIFAR10 and CIFAR100.

To control experimental variables, the same data augmentation techniques (random cropping and flipping) were applied to the training dataset. Convolutional layer parameters were initialized using Kaiming normal distribution, and the Momentum optimizer with an initial learning rate of 0.1 was used, along with cosine annealing for learning rate scheduling. Additionally, the evaluation metrics calculation employed the same sampling approach, with ten samples randomly selected for each label, amounting to ten times the number of training samples' labels.
\begin{figure}[tb]
	\centering
	\includegraphics[width=0.45\textwidth, keepaspectratio]{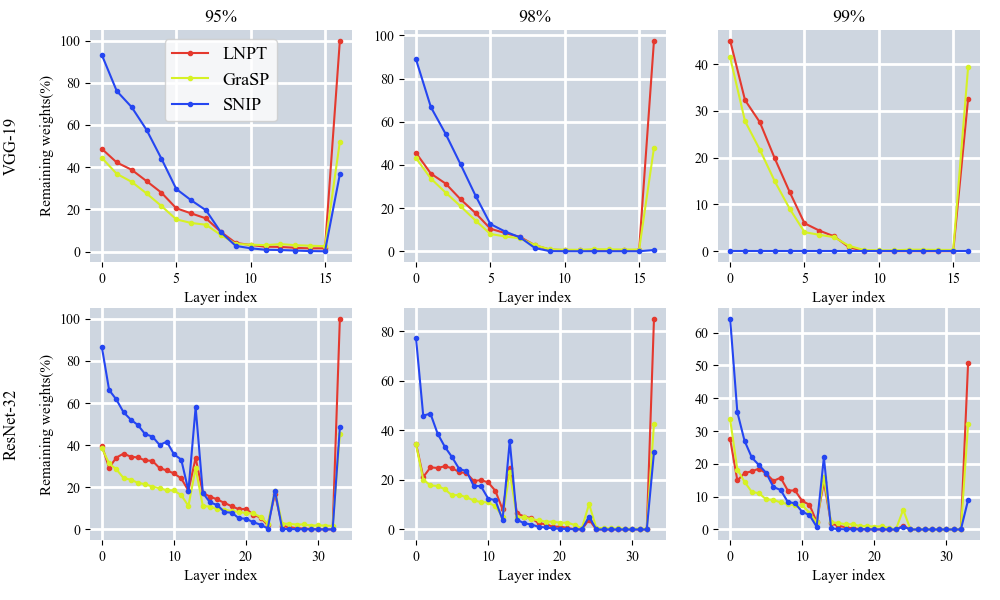}
	\caption{Pruning results for VGG-19 and ResNet-32 at different pruning rates on the CIFAR-10 dataset.}
	\label{fig4}
	\vskip 1pt
\end{figure}

The performance comparison of CIFAR datasets on VGG-19 and ResNet-32 is shown in Table \ref{table1} and Fig. \ref{fig3}. LNPT outperforms SNIP and GraSP in most cases. SNIP approaches the baseline at lower compression rates but falls short due to significant differences between teacher and student feature maps at high compression rates, where LNPT excels. For CIFAR-10 with fewer categories, LNPT achieves accuracies of 92.06\% and 85.63\% at 99\% pruning rate for VGG-19 and ResNet-32, respectively. For the more complex CIFAR-100 dataset, LNPT achieves accuracies of 67.33\% and 51.44\% for VGG-19 and ResNet-32, respectively. ResNet's compact structure and residual connections accentuate differences in feature maps, further showcasing LNPT's advantages. Notably, the combination of one-hot loss and intermediate semantics is crucial for training, as seen in previous work \cite{ref36}\cite{ref37}.

To further assess the effectiveness of our approach, we conducted experiments on the Tiny-Imagenet dataset. Table \ref{table2} reports the classification results of state-of-the-art pruning methods and LNPT on the Tiny-Imagenet dataset for the student network. At high compression rates, especially at a 99\% compression rate,, LNPT outperforms other methods significantly, achieving 53.21\% and 27.35\% accuracy on VGG-19 and ResNet-32, respectively.

Differences between LNPT and other pruning methods in terms of the proportion of remaining weights in sparse networks for each layer were investigated. As depicted in Fig.\ref{fig4}, it was observed that these pruning methods resulted in different pruning strategies.

LNPT focuses on gradient characteristics of feature maps, being more sensitive to tail-end weights and pruning more in those areas. However, at high pruning rates, the top layers of the network become smaller, and LNPT retains more weights in the top layers compared to GraSP. This indicates the significance of top weights in feature extraction, as they are highly sensitive to feature maps. Consequently, LNPT retains more top weights compared to GraSP, which prunes more top weights, potentially creating bottlenecks that impede feature information transmission.

Moreover, LNPT already had smaller $L_{\boldsymbol{m}}$ values compared to SNIP and GraSP at initialization. This indicates that our pruning method is highly effective in preserving weights sensitive to generalization from the start, further enhancing the student network's generalization capability. Unlike the approach in \cite{ref38}, our method only requires minimizing the feature map error in the penultimate layer without the need to optimize the error for the entire network.
\begin{table}[tb]
	\centering
	\caption{THE STRUCTURED PRUNING RESULTS ON CIFAR-10. 'PR' DENOTE PRUNING RATIO, 'PARA(PR)' DENOTES PARAMETERS PRUNING RATIO, 'BL' DENOTE 'BASELINE'.}
	\label{table3}
	\resizebox{1.0\columnwidth}{!}{
		\begin{tabular}{l c c c}
			\cmidrule[\lightrulewidth](lr){1-4}
			\textbf{Model} & \multicolumn{1}{c}{FLOPs(PR)} &
			\multicolumn{1}{c}{Para(PR)}&\multicolumn{1}{c}{Accuracy} \\ 
			\cmidrule[\lightrulewidth](lr){1-4}
			VGG-16(BL) &313.75M(0.0\%) & 14.72M(0.0\%) & 93.83\% \\
			L1\cite{ref45}	&206.00M(34.3\%)&	5.40M(63.3\%)&	93.40\%\\
			Zhao et al.\cite{ref46}	&190.00M(39.4\%)	&3.92M(73.3\%)	&93.18\%\\
			GAL-0.05\cite{ref40}	&189.49M(39.6\%)	&3.36M(77.1\%)	&93.03\%\\
			SSS\cite{ref39}	&183.16M(41.6\%)&	3.93M(73.3\%)&	93.02\%\\
			GAL-0.1\cite{ref40}	&171.89M(45.2\%)	&2.67M(81.8\%)	&90.73\%\\
			HRank\cite{ref15}	&108.61M(65.3\%)	&2.51M(82.9\%)	&92.34\%\\
			\textbf{Ours}	&100.71M(67.9\%)	&3.51M(76.1\%)	&92.57\%\\
			\textbf{Ours}	&47.6M(84.8\%)	&1.47M(90.0\%)	&91.45\%\\
			\cmidrule[\lightrulewidth](lr){1-4}
			ResNet-56(BL)&	125.49M(0.0\%)&	0.85M(0.0\%)	&93.26\%\\
			L1\cite{ref45}	&90.90M(27.6\%)&	0.73M(14.1\%)&	93.06\%\\
			HRank\cite{ref15}	&88.72M(29.3\%)	&0.71M(16.8\%)	&93.52\%\\
			NIPS\cite{ref47}	&81.00M(35.5\%)&	0.49M(42.4\%)&	93.01\%\\
			GAL-0.6\cite{ref40}	&78.30M(37.6\%)&	0.75M(11.8\%)&	92.98\%\\
			RNI\cite{ref52} & 71.86M(43.83\%)& - & 93.01\%\\
			\textbf{Ours}	&67.26M(45.4\%)&	0.46M(45.2\%)&	93.50\%\\
			HRank\cite{ref15}&	62.72M(50.0\%)&	0.49M(42.4\%)&	93.17\%\\
			
			\textbf{Ours}	&57.22M(54.4\%)	&0.33M(60.1\%)&	92.16\%\\
			GAL-0.8\cite{ref40}	&49.99M(60.2\%)	&0.29M(65.9\%)	&90.36\%\\
			\textbf{Ours}	&45.05M(64.1\%)	&0.22M(74.1\%)	&91.13\%\\
			\midrule
	\end{tabular}}
\end{table}
\begin{table}[tb]
	\centering
	\caption{THE STRUCTURED PRUNING RESULTS ON CIFAR-100.}
	\label{table4}
	\resizebox{1.0\columnwidth}{!}{
		\begin{tabular}{l c c c}
			\cmidrule[\lightrulewidth](lr){1-4}
			\textbf{Model} & \multicolumn{1}{c}{FLOPs(PR)} &
			\multicolumn{1}{c}{Para(PR)}&\multicolumn{1}{c}{Accuracy} \\ 
			\cmidrule[\lightrulewidth](lr){1-4}
			VGG-16(BL) &313.75M(0.0\%)	&14.72M(0.0\%)&	73.54\%\\
			Zhao et al.\cite{ref46}	&256.00M(18.4\%)&	9.14M(38.1\%)&	73.33\%\\
			DeepHoyer\cite{ref54}&211.30M(32.77\%)&-&71.66\%\\
			Polarization\cite{ref53}&200.59M(36.18\%)&-&71.23\%\\
			\textbf{Ours}	&196.09M(37.5\%)	&8.86M(39.8\%)&	73.44\%\\
			\cmidrule[\lightrulewidth](lr){1-4}
			ResNet-56(BL)&	125.49M(0.0\%)&	0.85M(0.0\%)	&73.09\%\\
			\textbf{Ours}	&71.27M(43.2\%)	&0.54M(35.3\%)	&72.82\%\\
			\textbf{Ours}	&60.10M(52.1\%)	&0.37M(55.3\%)&	70.55\%\\
			QSFM-SSIM\cite{ref48}	&58.38M(53.8\%)	&0.42M(51.1\%)	&68.36\%\\
			QSFM-PSNR\cite{ref48}	&58.38M(53.8\%)	&0.42M(51.1\%)	&68.33\%\\
			\midrule
	\end{tabular}}
\end{table}
\subsection{Structured Pruning}
\noindent Two network architectures, VGG-16 and ResNet-56, were used, and the rest of the training details were kept similar to non-structured pruning. During the pruning phase, we accumulated the weight scores within channels to obtain the channel score. \cite{ref25} specified the pruning ratio for each layer at the beginning of training, while LNPT only required specifying a global pruning ratio, and LNPT did not require repeated pruning processes. Our method is single-shot, fast, and achieves better results.

For the CIFAR-10 dataset, the experimental results are shown in Table \ref{table3}. It can be seen that whether using VGG-16 or ResNet-56, our method can effectively reduce FLOPs and parameters, even suppressing overfitting effects and improving accuracy on certain networks.

For VGG-16, LNPT reduces FLOPs by 67.9\%, while maintaining high performance in the pruned network. 90\% of parameters were pruned, yet network performance remained high. In the case of ResNet-56, almost half of parameters (45.2\%) were pruned, resulting in a 0.24\% accuracy increase. With a 64.1\% FLOPs reduction, the pruned network achieves 91.13\% accuracy, with pruned parameters comprising only 25.9\%.

Table \ref{table4} shows pruning results for various network structures on CIFAR-100 dataset. Compared to CIFAR-10, CIFAR-100 has more classes (100 classes), making it more challenging and likely to decrease network performance. However, LNPT still performs well. Taking VGG16 and ResNet-56 as examples, FLOPs reduce by 37.5\% and 43.2\%, respectively, with only a 0.1\% and 0.27\% decrease in accuracy.

\subsection{Analysis}
\begin{figure}[tb]
	\centering
	\includegraphics[width=0.45\textwidth, keepaspectratio]{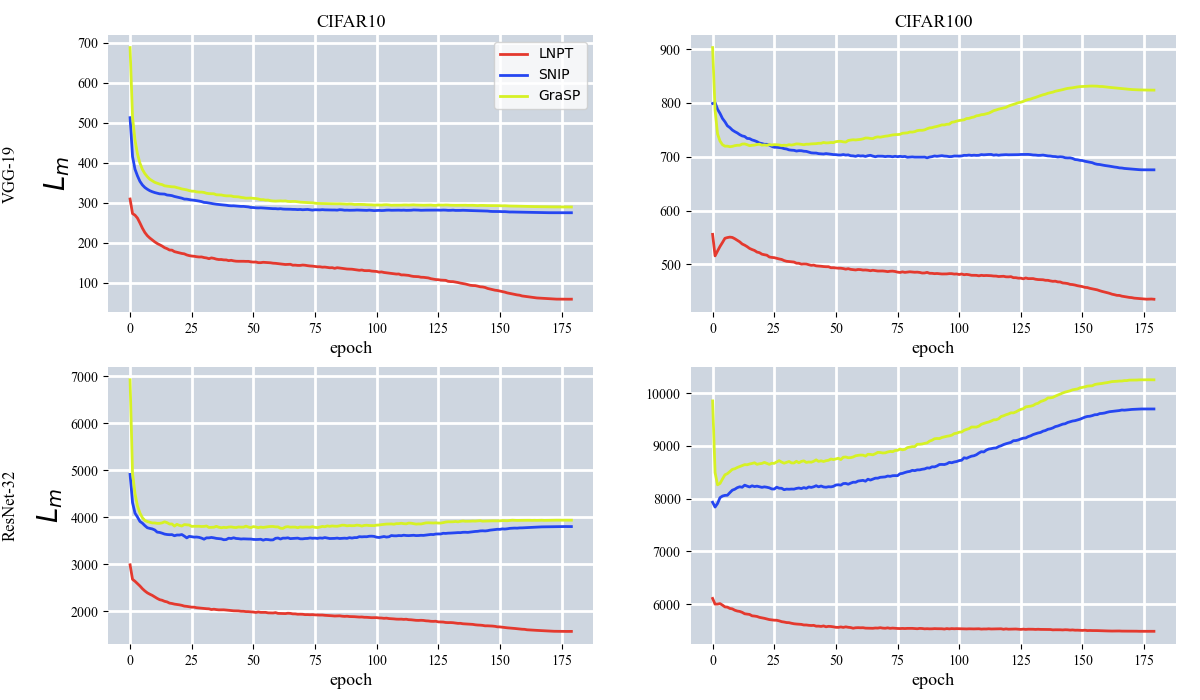}
	\caption{The change in $L_{\boldsymbol{m}}$ under different pruning algorithms at a 95\% compression rate. The sampling method involves summing $L_{\boldsymbol{m}}$ across all data points and calculating the mean value.}
	\label{fig5}
\end{figure}

\noindent \textbf{Learning gap.} To further evaluate the performance of our proposed optimization objective, we examined the changes in $L_{\boldsymbol{m}}$ under different pruning algorithms, as shown in Fig.\ref{fig5}. LNPT consistently achieved the lowest $L_{\boldsymbol{m}}$ values in all cases, and these values continued to decrease throughout the training process, eventually converging to a certain value, similar to the generalization loss.

\begin{figure}[tb]
	\centering
	\includegraphics[width=0.4\textwidth, keepaspectratio]{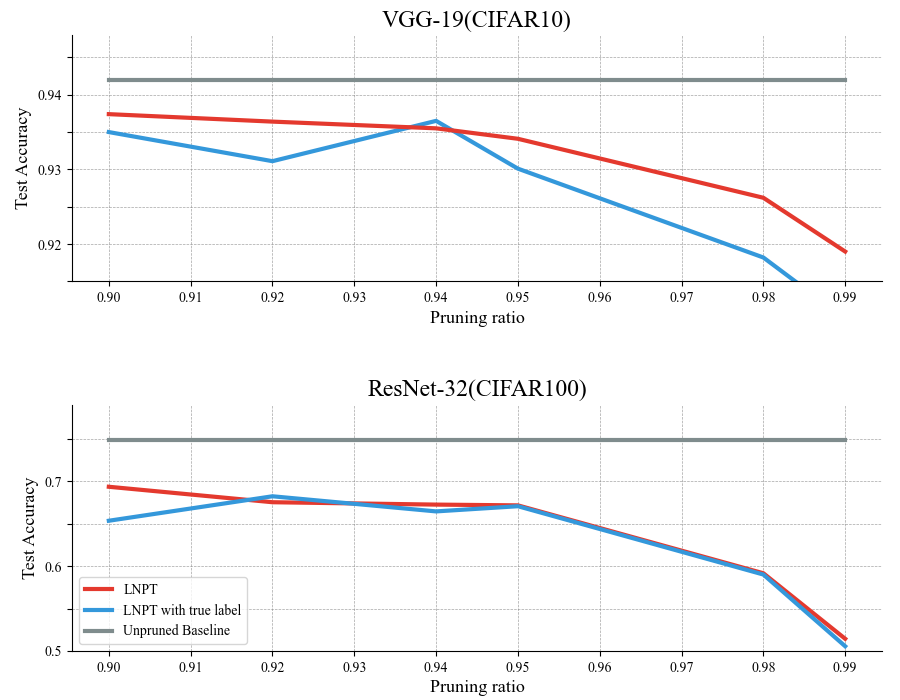}
	\caption{The performance of LNPT and LNPT using true labels at different pruning ratios}
	\label{fig6}
\end{figure}

\textbf{LNPT with true labels.} We introduced a training approach for the student model using both pseudo-labels and feature maps. To assess the impact of this factor, we conducted an ablation experiment where the model was trained exclusively using real label losses. The results, as depicted in Fig. \ref{fig6}, clearly demonstrate that LNPT continues to outperform, underscoring the effectiveness of the combination of pseudo-labels and feature maps in endowing the student model with richer knowledge and achieving superior performance compared to using only real labels.

\subsection{An empirical approach for $\boldsymbol{C}^t$}
\noindent In \ref{section method}, we qualitatively measure the generalization performance of the pruned network by using $\Delta \boldsymbol{m}$ to represent $\boldsymbol{\ell}$. In this section, an ideal approximation for $\boldsymbol{C}^t$ is provided, which is a focus of our future work. At time t, the change in $\Delta \boldsymbol{m}$ for the fitting target $Loss\left(\boldsymbol{\theta}_s^t\right)$ is represented as follows:
\begin{equation}
	\label{eq16}
	\Delta\boldsymbol{\ell}=\Delta \boldsymbol{m}^{t+1}-\Delta \boldsymbol{m}^t=\eta\nabla_{\boldsymbol{\theta}_s^t}\boldsymbol{m}_s^t\nabla_{\boldsymbol{\theta}_s^t}Loss\left(\boldsymbol{\theta}_s^t\right)
\end{equation}

At this point, combined with (\ref{eq3}), we can explicitly consider $\boldsymbol{C}^t$ as $\nabla_{\boldsymbol{\theta}_s^t}\boldsymbol{m}_s^t$.

As shown in Fig.\ref{fig5}, our proposed $L_{\boldsymbol{m}}$ steadily decreases, accurately capturing the network's generalization dynamics.

$\boldsymbol{C}^t$ can provide a reasonable explanation for training before pruning. Previously, we aimed to preserve parameters sensitive to $\Delta\boldsymbol{d}$. However, since $\Delta\boldsymbol{d}$ cannot accurately reflect the network's generalization ability over the long training process, pruning methods based on $\Delta\boldsymbol{d}$ cannot identify which parameters are globally effective. In the case of $\Delta\boldsymbol{\ell}$, since $\boldsymbol{\ell}$ steadily decreases throughout the training process, by preserving parameters sensitive to $\Delta\boldsymbol{\ell}$, parameters that are effective for generalization on a global scale can be identified.

\section{CONCLUSION}
In this paper, common perspectives on the generalization properties of sparse networks were evaluated, and the inappropriateness of these metrics during training was explained. Our proposed learning gap addresses this inappropriateness and provides a reasonable explanation for pre-pruning. We hope that this will guide further understanding of sparse learning and deep learning theory. Additionally, based on the learning gap, LNPT was introduced for adaptive pruning and training of networks without training labels. Experiments on various datasets have shown that this method can adaptively prune and train the student based on the teacher hosted on the server or in the cloud without requiring data labels.


\bibliographystyle{IEEEtran}

\bibliography{bibtex}

\end{document}